%% file: cpf.tex
\documentclass[letterpaper, 10 pt, conference]{ieeeconf}

\usepackage{mathtools}
\usepackage[pdftex,final]{graphicx}
\usepackage{units}
\usepackage[english]{babel}
\usepackage[latin1]{inputenc}
\usepackage[usenames]{color}
\usepackage{comment}
\usepackage{balance}
\usepackage{url}
\usepackage{todonotes}
\usepackage{amssymb}
\usepackage{tikz}

\usetikzlibrary{bayesnet}

\makeatletter
\let\NAT@parse\undefined
\makeatother
\usepackage[numbers]{natbib}

\usepackage{xcolor}
\usepackage{textpos}
\usepackage{setspace}
\usepackage[bookmarks=true]{hyperref}
\usepackage[all]{hypcap}

\hypersetup{
    colorlinks,
    linkcolor={red!50!black},
    citecolor={blue!50!black},
    urlcolor={blue!80!black}
}

\makeatletter
\def\input@path{{../figures/}}
\makeatother
\graphicspath{{../figures/}}

\IEEEoverridecommandlockouts                             
\newcommand{\sta}{x} 
\newcommand{\obs}{y} 
\newcommand{\pno}{v} 

\newcommand{\sam}{\varphi} 

\newcommand{\pre}{0}
\newcommand{\cur}{1}
\newcommand{\nex}{2}

\title{\LARGE \bf
The Coordinate Particle Filter - \\A novel Particle Filter for High Dimensional Systems
}

\author{Manuel W\"uthrich$^{1}$, Jeannette Bohg$^{1}$, Daniel
  Kappler$^{1}$, Claudia Pfreundt$^{2}$ and Stefan Schaal$^{1,3}$ 
  \thanks{$^{1}$ Autonomous Motion Department at the Max-Planck-Institute for Intelligent Systems, T\"ubingen, Germany
    Email: {\tt\small   first.lastname@tuebingen.mpg.de}}%
  \thanks{$^{2}$ High Performance Humanoid Technologies Lab, Karlsruhe
    Institute of Technology, Karlsruhe, Germany}%
  \thanks{$^{3}$ Computational Learning and Motor Control lab at the
  University of Southern California, Los Angeles, CA, USA}%
}

\begin{document}

\maketitle
\thispagestyle{empty}
\pagestyle{empty}

\begin{abstract}
Parametric filters, such as the Extended Kalman Filter and the Unscented Kalman Filter, 
typically scale well with the dimensionality of the problem, but they are known to fail if
the posterior state distribution cannot be closely approximated 
by a density of the assumed parametric form.

For nonparametric filters, such as the Particle Filter, the converse holds. 
Such methods are able to approximate any posterior, but the computational requirements scale exponentially
with the number of dimensions of the state space.
In this paper, we present the Coordinate Particle Filter which alleviates this problem. 
We propose to compute the particle weights recursively, dimension by dimension. 
This allows us to explore one dimension at a time, and resample after each dimension if necessary.

Experimental results on simulated as well as real data confirm that the proposed method
has a substantial performance advantage over the Particle Filter in high-dimensional systems where
not all dimensions are highly correlated. We demonstrate the benefits 
of the proposed method for the problem of multi-object and robotic manipulator tracking.
\end{abstract}

\begin{textblock*}{100mm}(.\textwidth,-12cm)
 \begin{spacing}{0.8}
 {\fontsize{8pt}{2pt}\selectfont \sffamily
\noindent 2015 IEEE International Conference on\\
Robotics and Automation (ICRA)\\
May 26-30, 2015. Seattle, USA}
\end{spacing}
\end{textblock*}%

\section{Introduction}\label{sect:introduction}
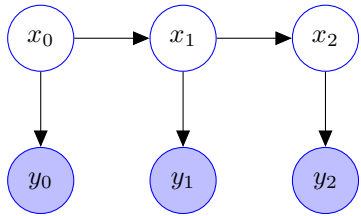
\begin{figure}[h]
  \begin{center}
      \input{hidden_markov} 
  \end{center}
  \caption{The belief network which characterizes the evolution of the state $x$ and the observations $y$.}
  \label{fig:hidden_markov}
\end{figure}

Decision making requires knowledge of some variables of interest. In the vast majority of real-world problems these variables are latent, i.e. they cannot be observed directly but have to be inferred based on sensor measurements. If decision making has to be performed online, these latent variables have to be inferred online as well. Therefore, past measurements have to be fused with incoming measurements to always maintain an up-to-date belief over the latent variables. This problem is called filtering and the underlying dynamical system is typically modeled with a State Space Model (see Fig.~\ref{fig:hidden_markov}).

More formally, filtering means inferring the state $\sta$ at a given time, knowing only the measurements $\obs$ up to that time. Applications range from robotics, over estimating a digital communication signal using noisy measurements, to estimating the volatility of financial instruments using stock market data \cite{doucet_book}. Often these systems have high dimensional state spaces. For example, the full joint state of a humanoid robot is typically over 50 dimensional. One of the most popular algorithms for performing inference in dynamical systems, the {\em Particle Filter\/} (PF), breaks down in such high dimensional systems \cite{li, bickel}. In this paper, we address this issue and propose a novel filter called the {\em Coordinate Particle Filter\/} (CPF) which scales well with the dimensionality of the state for systems where only a subset of the state dimensions is highly correlated.

\subsection*{ Notation and Problem Statement}
We assume the underlying process of the dynamical system to be stationary. Therefore the absolute time indices are irrelevant, only the time difference matters within a figure or equation. To simplify notation we will thus use the indices $0,1,2$ throughout the paper. Furthermore, we introduce the notation $:t$ to denote all the time steps up to $t$. 

The system is described by two functions, the process model $g$ and the measurement model $h$.
\begin{align}
 x_{ 2 }&=g(x_{ 1 },v_{ 2 }) \label{eq:process_model}\\ 
 y_{ 2 }&=h(x_{ 2 },w_{ 2 }) \label{eq:measurement_model}
\end{align}
Without loss of generality, the noise variables $v$ and $w$ can be modeled as normally distributed with zero mean and unit variance, since they can always be mapped onto different distributions inside of the process and measurement models.

In filtering, the distribution of interest is the current belief given all the measurements taken so far $p(x_{ 2 }|y_{ :2 })$. It can be computed recursively as follows
\begin{align}
p(x_{ 2 }|y_{ :2 })\propto p(y_{ 2 }|x_{ 2 })\int _{ \chi } p(x_{ 2 }|x_{ 1 }=\chi)p(x_{ 1 }=\chi|y_{ :1 })\label{eq:bayes_filter}
\end{align}
which is the well known Bayes filter. Only in very few cases can this integral be solved analytically, there are thus numerous approximation methods, the most prominent ones being the Extended Kalman Filter (EKF), the Unscented Kalman Filter (UKF) and the Particle Filter.

The EKF is known to fail if the system exhibits substantial non-linearity \cite{cappe_overview}. Many algorithms have been proposed to improve the performance of the EKF, most prominently the Unscented Kalman Filter (UKF) \cite{ukf}. The UKF has been applied successfully in many settings where the posterior distribution can be closely approximated by a Gaussian. In nonlinear systems this assumption can be violated, which often precludes the usage of a UKF \cite{cappe_overview}. 

These limitations have led to a big interest in alternative methods which can represent a bigger family of dynamical systems. Sequential Monte Carlo, i.e. Particle Filtering methods, started to be used widely since the seminal work by \citet{gordon}. These methods are applicable in general state-space models and allow the computation of all kinds of moments, quantiles and highest posterior density regions \cite{cappe_overview}, whilst the EKF and UKF approximate only the first and second order moments. The Particle Filter has found applications in 
practically all areas of signal processing concerned with Bayesian dynamical models \cite{cappe_overview}, such as signal processing, control, robotics, finance, and statistics.

Unfortunately the PF is not well suited for high dimensional systems. The number of particles $N$ has to scale exponentially with the number of dimensions $D$ in order to prevent the filter from failing \cite{li, bickel}.
We present a method to alleviate this problem in Section~\ref{sect:coordinate_particle_filter}.

\section{Monte Carlo and the Curse of Dimensionality}
In this section we will briefly review some basic Monte Carlo methods and subsequently explain the Particle Filter, for a more extensive discussion we refer the reader to \cite{gordon, cappe_overview, barber, robert}. The objective in Particle Filtering is to approximate expectations with respect to the posterior from Eq.~\ref{eq:bayes_filter}, $p(x_2|y_{:2})$, using a set of samples $\{\chi_2^l\}$. If we draw a set of samples $\{\chi_{:2}^l\} \sim p(x_{:2}|y_{:2})$ we can always simply discard $\{\chi_{:1}^l\}$ in order to obtain a set of samples $\{\chi_2^l\} \sim p(x_{2}|y_{:2})$. There is therefore no need to integrate out previous states and Eq.~\ref{eq:bayes_filter} becomes
\begin{align}
 p(x_{ :2 }|y_{ :2 })\propto  p(y_{ 2 }|x_{ 2 })p(x_{ 2 }|x_{ 1 })p(x_{ :1 }|y_{ :1 }).\label{eq:full_recursion}
\end{align}
We will thus be concerned with drawing, or approximately drawing samples $\{\chi_{:2}^l\} \sim p(x_{:2}|y_{:2})$.

Firstly we will recall some basic notions of sampling which are needed to follow the subsequent derivations. For more details, please refer to \cite{owen, robert, barber}.

\subsection{Simple Monte Carlo}
We are interested in approximating expectations using sampling. That is, we want to find a set of samples $\{\chi^l\}$, such that
\begin{align}
\int _{\chi} f(\chi)p(x=\chi) \approx  \frac{1}{L}\sum _{l=1}^L f(\chi^l) \label{eq:simple_monte_carlo}
\end{align}
i.e. the right hand side is a good approximation for the left hand side.

It is easy to show that, if the samples $\{\chi^l\}$ are drawn from $p(x)$, then Eq.~\ref{eq:simple_monte_carlo} provides a consistent estimator, which means that for the number of samples $L\to \infty$ the two sides become equal \cite{owen, barber}. Therefore, given enough samples, we can compute the mean, the covariance or any other property of the distribution $p(x)$. The standard deviation of the estimate is proportional to $\frac{1}{\sqrt{L}}$, independently of the dimension.

The discussion above implies that, if we were able to sample efficiently from the posterior from Eq.~\ref{eq:full_recursion}, $p(x_{:2}|y_{:2})$, then we could approximate the expectations of interest even in high dimensional systems with no need for an excessive number of samples. Unfortunately, in the vast majority of the cases of interest, this distribution is highly complex and it is impossible to sample from it directly \cite{owen, barber, doucet_book}. Therefore more sophisticated schemes have to be used to evaluate expectations in most real world problems.

\subsection{Importance Sampling}
Importance Sampling (IS) is such a scheme which can be used when it is impossible to sample from $p(x)$. Since it is not possible to generate samples from the random variable $x$ we introduce an auxiliary random variable $\varphi$ which is distributed according to the so called proposal distribution. This distribution is chosen such that we can easily draw samples from it. To use these samples for computing expectations with respect to $x$, they have to be weighted to account for the mismatch between the distributions of $x$ and $\varphi$.

To find these weights we expand Eq.~\ref{eq:simple_monte_carlo} as 
\begin{align*}
\int _{ \chi } f(\chi)p(x=\chi)&=\frac { \int _{ \chi } f(\chi)p(x=\chi) }{ \int _{ \chi } p(x=\chi) } \\ 
&=\frac { \int _{ \chi } f(\chi)\frac { p(x=\chi) }{ p(\varphi = \chi) } p(\varphi = \chi) }{ \int _{ \chi } \frac { p(x = \chi) }{ p(\varphi = \chi) } p(\varphi = \chi) } \\ 
&=\frac { \int _{ \chi } f(\chi)\omega (\chi)p(\varphi = \chi) }{ \int _{ \chi } \omega (\chi)p(\varphi = \chi) } 
\end{align*}
where we have defined $\omega (\chi)\propto\frac { p(x = \chi) }{ p(\varphi = \chi) } $. Very importantly, $\omega (\chi)$ can be any function which is proportional to $\frac { p(x = \chi) }{ p(\varphi = \chi) } $, since it appears in the numerator as well as the denominator and any constant will canceled. This means that we need to know $p(x)$ and $p(\varphi)$ only up to a normalization constant. The distribution $p(\varphi)$ can be chosen freely, as long as its support contains the support of $p(x)$. Choosing $p(\varphi)$ such that we can sample from it, we can apply simple Monte Carlo to approximate both the numerator and the denominator separately
\begin{align*}
 \int _{ \chi } f(\chi)p(x=\chi)\approx \frac { \sum _{ l=1 }^L f(\chi^{ l })\omega (\chi^{ l }) }{ \sum _{ l=1 }^L \omega (\chi^{ l }) } 
\end{align*}
with $\{\chi^l\}\sim p(\varphi)$. It is easy to show that this estimator converges to the true expectation with probability equal to $1$ as $L \to \infty$ \cite{owen}. The standard deviation of the estimator is again proportional to $\frac{1}{\sqrt{L}}$, but, as opposed to simple Monte Carlo, it now also depends heavily on the proposal distribution $p(\varphi)$. There will necessarily be some mismatch between the proposal and the desired distribution, which leads to the variance of the estimator growing exponentially in the number of dimensions of the state space \cite{barber, owen}. This high variance is a consequence of weight degeneracy, i.e. almost all of the weight being concentrated on a very small subset of the samples.

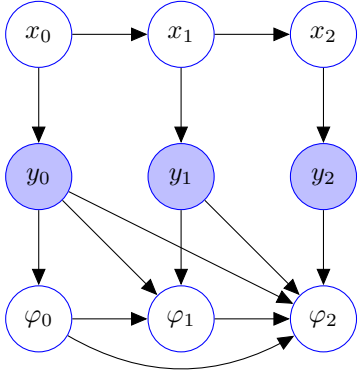
\begin{figure}[t]
  \begin{center}
      \input{standard_particle_filter} 
  \end{center}
  \caption{The belief network describing the evolution of the random process of interest $x$ and another artificial random process $\varphi$.}
  \label{fig:standard_particle_filter}
\end{figure}

Let us now apply IS to the problem we are trying to address, which is computing expectations with respect to $ p(x_{ :2 }|y_{ :2 })$. Since we cannot sample from this distribution directly, we define an alternative process $p(\varphi_{ :2 }|y_{ :2 })$ which approximates the actual process, see Fig.~\ref{fig:standard_particle_filter}.
Having chosen the proposal distribution we have to compute the weights in order to account for the mismatch between the distribution of interest and the proposal distribution.
\begin{align}
\omega _{ 2 }(\chi _{ :2 })\propto\frac { p(x_{ :2 }=\chi _{ :2 }|y_{ :2 }) }{ p(\varphi _{ :2 }=\chi _{ :2 }|y_{ :2 }) }\label{eq:batch_weight}
\end{align}
In dynamical systems, these weights can be computed sequentially, which is called Sequential Importance Sampling (SIS).
\subsection{Sequential Importance Sampling}\label{sec:sis}
Fig.~\ref{fig:standard_particle_filter} shows both the process $x$ we are trying to estimate and the auxiliary process $\varphi$ which we created. When sampling, we have knowledge of all the past observations and the past samples we generated, therefore $\varphi_2$ can depend on all the preceding $\varphi$ and $y$.

To compute the weights sequentially, we have to write both the denominator and the numerator of Eq.~\ref{eq:batch_weight} recursively. The numerator can be computed recursively according to Eq.~\ref{eq:full_recursion}. The denominator, i.e. the proposal distribution, can be written recursively as well. Using the independence assumptions from the belief network from Fig.~\ref{fig:standard_particle_filter} we obtain
\begin{align}
 p(\varphi _{ :2 }|y_{ :2 })=p(\varphi _{ 2 }|\varphi _{ :1 },y_{ :2 })p(\varphi _{ :1 }|y_{ :1 }). \label{eq:sampling_dynamics}
\end{align}
Using this result and Eq.~\ref{eq:full_recursion}, Eq.~\ref{eq:batch_weight} can be re-written as:
\begin{align}
  \omega_2 (\chi_{:2}) \propto\frac { p(y_{ 2 }|x_{ 2 }\!=\!\chi _{ 2 })p(x_{ 2 }=\chi _{ 2 }|x_{ 1 }=\chi _{ 1 }) }{ p(\varphi _{ 2 }=\chi _{ 2 }|\varphi _{ :1 }=\chi _{ :1 },y_{ :2 }) } \omega _{ 1 }(\chi _{ :1 }). \label{eq:seq_weight}
\end{align}
The proposal distribution can be chosen freely, a very common choice is $p(\varphi _{ 2 }=\chi _{ 2 }|\varphi _{ :1 }=\chi _{ :1 },y_{ :2 })=p(x_{ 2 }=\chi _{ 2 }|x_{ 1 }=\chi _{ 1 })$, the above recursion then simplifies to 
\begin{align*}
 \omega _{ 2 }(\chi _{ :2 })\propto p(y_{ 2 }|x_{ 2 }=\chi _{ 2 })\omega _{ 1 }(\chi _{ :1 }).
\end{align*}
Resuming, SIS provides a way of computing the weights recursively. The final weights will however be exactly the same as in standard IS. SIS thus suffers from weight degeneracy in the exact same way as IS does. As discussed above, due to the unavoidable mismatch between proposal distribution and distribution of interest, the number of particles has to grow exponentially in the number of dimensions. The dimension of the sample $\chi _{ :2 }$ is equal to $DT$ where $T$ is the number of time steps and $D$ is the dimension of the state. Since $DT$ is typically very high and growing in time, computing with an exponential number of particles is intractable, and weight degeneracy thus unavoidable. The solution proposed in \cite{gordon} is to discard particles with low weights, which will not contribute significantly to the computation of the expectation anyways, by resampling. This algorithm is well known as the Particle Filter.

\subsection{The Particle Filter}
In Particle Filtering, we use the standard SIS algorithm as long as the weights are not degenerate. As soon as they become too
concentrated, according some criterion, we resample.
The most common way of resampling is to redraw each particle with a
probability proportional to its weight.
Any resampling strategy which does not bias the estimator
can be used \cite{cappe_overview}.

Since in a Particle Filter we can resample after each time step, the exploration space has thus effectively been reduced from $TD$ to $D$ dimensions.

If $D$ is small enough such that the state space can be explored reasonably well
by $N$ samples, the problem of weight degeneracy is resolved.
In many systems however, the dimension of the state space $D$ is too large large to be covered by any
tractable number of particles.
Therefore, Particle Filtering in high dimensional state spaces
remains an open problem \cite{li,bickel,owen}.

\section{Proposed Method}\label{sect:coordinate_particle_filter}
The key idea of the Particle Filter is to compute the weights recursively in time, such that it is possible to resample after each time step. Here we extend this idea and propose to compute the weights not just recursively in time, but also in the dimensions of the state space, which will allow for a resampling step after each dimension. Hence, the dimensionality of the space explored in one step is reduced from $TD$ to $D$ through the usage of a standard Particle Filter and we propose to further reduce it to $1$. The presented method is referred to as the Coordinate Particle Filter (CPF), since it is reminiscent of Coordinate Descent.

More concretely, we will inject the noise $v_2$ dimension by dimension into the process described by Eq.~\ref{eq:process_model} and update the weights after each step. We are therefore required to write the weights recursively in the dimensions of the noise.

\subsection{Explicit noise}
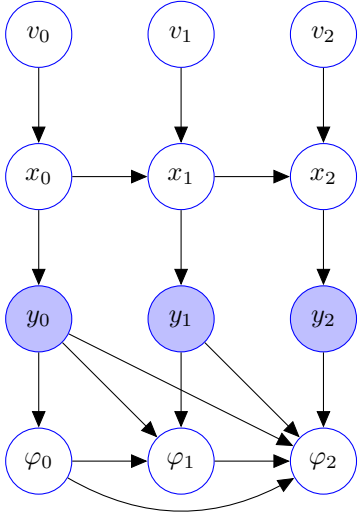
\begin{figure}[h]
  \begin{center}
      \input{noise_particle_filter} 
  \end{center}
  \caption{This is the same system as in Fig.\ref{fig:standard_particle_filter}, with explicit process noise $v$.
  The measurement noise $w$ is still implicit.}
  \label{fig:noise_particle_filter}
\end{figure}
To facilitate the subsequent derivation of the proposed method, we formulate the standard Particle Filter in terms of the noise variables $v$. Making the process noise explicit in Fig.~\ref{fig:standard_particle_filter}, we obtain Fig.~\ref{fig:noise_particle_filter}. Since the only source of uncertainty in the state trajectory is the process noise $v$, knowledge of the noise trajectory $v_{:2}$ implies knowledge of the current state $x_{2}$. 
%
%
Therefore $v_{:2}$ is a valid representation of the state $x_{2}$ of the dynamical system. In practice, it would of course be very inefficient to store the state as the noise trajectory, but conceptually we can use $v_{:2}$ and $x_{2}$ interchangeably. Applying this substitution to Eq.~\ref{eq:full_recursion} we obtain
\begin{align}
p(v_{ :2 }|y_{ :2 }) &\propto p(y_{ 2 }|v_{ :2 })p(v_{ 2 })p(v_{ :1 }|y_{ :1 }) \label{eq:noise_dynamics}
\end{align}
where we have made use of the fact that the process noises at different time steps are independent. Similarly, we can write the weights in Eq.~\ref{eq:seq_weight} in terms of the noise variables.
\begin{align}
 \omega _{ 2 }(\nu _{ :2 })\propto \frac { p(y_{ 2 }|v_{ :2 }=\nu _{ :2 })p(v_{ 2 }=\nu _{ 2 }) }{ p(\varphi _{ 2 }=\nu _{ 2 }|\varphi _{ :1 }=\nu _{ :1 },y_{ :2 }) } \omega _{ 1 }(\nu _{ :1 }) \label{eq:seq_noise_weights}
\end{align}
This somewhat unusual formulation is equivalent to the standard Particle Filter and it will provide a basis for the proposed extension.
\subsection{Computing the weights recursively in the dimension}
The above recursion has the form $\omega _{ 2 }(\nu _{ :2 })\propto f(\nu _{ :2 })\omega _{ 1 }(\nu _{ :1 })$, where we omit the dependency on $y$ for simplicity. The current weights are obtained by multiplying the previous weights with some function of the extended noise trajectory, subsequently they can be resampled if necessary for preventing weight degeneracy. However, if the dimension of the problem is very large, the weights can already be degenerate after just one time step.

We therefore go one step further and write the weights recursively in the dimensions $d$ as well. More precisely, we want to find the weight of the noise trajectory $\{\nu_{:1}, \nu_{2}^{1:d} \}$ up until dimension $d$ as a function of the weight of the noise trajectory $\{\nu_{:1}, \nu_{2}^{1:d-1} \}$ up until dimension $d-1$. To keep notation uncluttered we define  $\nu_{ :2 }^{ :d } = \{\nu_{:1}, \nu_{2}^{1:d} \}$, which denotes all the past noises $\nu_{:1}$ and the current noise $\nu_2^{1:d}$ up until dimension $d$. Furthermore it will be convenient to use the convention $\nu_{ :2 }^{ :0 }=\nu_{ :1 }^{:D}=\nu_{ :1 }$.
Using this notation we can write the objective of this section as finding a recursion of the form 
$\omega _{ 2 }^{ d }(\nu _{ :2 }^{ :d })\propto f(\nu _{ :2 }^{:d})\omega _{ 2 }^{ d-1 }(\nu _{ :2 }^{ :d-1 })$. 

Similarly to Eq.~\ref{eq:batch_weight}, the weights are defined as the ratio between the desired and the proposal distribution.
\begin{align}
 \omega _{ 2 }^d(\nu _{ :2 }^{ :d })\propto \frac { p(v_{ :2 }^{ :d }=\nu _{ :2 }^{ :d }|y_{ :2 }) }{ p(\varphi _{ :2 }^{ :d }=\nu _{ :2 }^{ :d }|y_{ :2 }) }\label{eq:noise_weights}
\end{align}
We will inject the noise at time $1$ dimension by dimension, then the noise at time $2$ dimension by dimension and so on. Two recursions are therefore required, one in time and one in the noise dimension. 

\subsubsection{Time Recursion}
The time recursion has to express the current weight $\omega_2^0$ as a function of the previous weight $\omega_1^D$. We use the independence assumptions implied by Fig.~\ref{fig:noise_particle_filter} to express both the numerator
\begin{align}
p(v_{ :2 }^{:0}|y_{ :2 })=p(v_{ :1 }^{:D}|y_{ :2 })\propto p(y_{ 2 }|v_{ :1 }^{:D})p(v_{ :1 }^{:D}|y_{ :1 })\label{eq:dim_process_dynamics_init}
\end{align}
and the denominator 
\begin{align}
  p(\varphi _{ :2 }^{:0}|y_{ :2 }) =  p(\varphi _{ :1 }^{:D}|y_{ :2 }) \propto p(\varphi _{ :1 }^{:D}|y_{ :1 }).\label{eq:dim_sampling_dynamics_init}
\end{align}
Finally, we insert these two equations into Eq.~\ref{eq:noise_weights} for dimension $d=0$ to obtain
\begin{align}
  \boxed
  {
 { \omega  } _{ 2 }^0(\nu _{ :2 }^{:0})\propto  { p } (y_{ 2 }|v_{ :1 }^{:D}=\nu _{ :1 }^{:D}) { \omega  } _{ 1 }^D(\nu _{ :1 }^{:D}).\label{eq:cpf_initial_weight_recursion}
  }
\end{align}
This equation defines the time recursion, it incorporates the measurement at time $2$, without extending the noise trajectory yet.

\subsubsection{Dimension Recursion}
To extend the noise trajectory we write the numerator and denominator of Eq.~\ref{eq:noise_weights} recursively in the dimension $d$.
Using the independence assumptions implied by the belief network in Fig.~\ref{fig:noise_particle_filter} we can rewrite Eq.~\ref{eq:noise_dynamics} to obtain a recursive expression for the numerator of Eq.~\ref{eq:noise_weights}
\begin{align}
p(v_{ :2 }^{ :d }|y_{ :2 })&\propto p(v_{ 2 }^{ d }|v_{ :2 }^{ :d-1 },y_{ 2 })p(v_{ :2 }^{ :d-1 }|y_{ :2 })\notag \notag \\ 
&\propto \frac { p(y_{ 2 }|v_{ :2 }^{ :d })p(v_{ 2 }^{ d }) }{ p(y_{ 2 }|v_{ :2 }^{ :d-1 }) } p(v_{ :2 }^{ :d-1 }|y_{ :2 })\label{eq:dim_process_dynamics}
\end{align}
Similarly we can rewrite the sampling dynamics from Eq.~\ref{eq:sampling_dynamics} to obtain a recursive expression for the denominator
\begin{align}
 p(\varphi _{ :2 }^{ :d }|y_{ :2 }) &\propto p(\varphi _{ 2 }^{ d }|\varphi _{ :2 }^{:d-1},y_{ :2 })p(\varphi _{ :2 }^{ :d-1 }|y_{ :2 }).\label{eq:dim_sampling_dynamics}
\end{align}
Inserting Eq.~\ref{eq:dim_process_dynamics} and Eq.~\ref{eq:dim_sampling_dynamics} into Eq.~\ref{eq:noise_weights} yields the equation for injecting the noise dimension by dimension
\begin{align}
   \boxed{
 { \omega  } _{ 2 }^d(\nu _{ :2 }^{ :d })\propto \frac {  { p } (y_{ 2 }|v_{ :2 }^{ :d }=\nu _{ :2 }^{ :d }) }{  { p } (y_{ 2 }|v_{ :2 }^{ :d-1 }\!\!=\nu _{ :2 }^{ :d-1 }) } \phi_1^d (\nu _{ :1 }^{ :d }) { \omega  } _{ 2 }^{d-1}(\nu _{ :2 }^{ :d-1 })\label{eq:cpf_weight_recursion}
   }
\end{align}
where we have defined 
\begin{align}
 \phi_1^d (\nu _{ :1 }^{ :d }):=\frac { p(v_{ 2 }^{ d }=\nu _{ 2 }^{ d }) }{ p(\varphi _{ 2 }^{ d }=\nu _{ 2 }^{ d }|\varphi _{ :2 }^{:d-1}=\nu _{ :2 }^{:d-1},y_{ :2 }) }. \notag
\end{align}

\subsubsection{Algorithm} \label{sec:algorithm}
Eq.~\ref{eq:cpf_initial_weight_recursion} and Eq.~\ref{eq:cpf_weight_recursion} enable us to update the weights dimension wise, as desired:
\begin{itemize}
\item We start out with the previous weight $ { \omega  } ^D_{ 1 }(\nu _{ :1 }^{:D})$.
\item We apply Eq.~\ref{eq:cpf_initial_weight_recursion} to obtain $ { \omega  } ^0_{ 2 }(\nu _{ :2 }^{:0})$.
\item We iteratively apply Eq.~\ref{eq:cpf_weight_recursion} to obtain $ { \omega  } ^D_{ 2 }(\nu _{ :2 }^{:D})$.
\end{itemize}
Writing out the above algorithm we obtain
\begin{align*}
 \omega _{ 2 }^D(\nu _{ :2 }^{:D})\propto \frac { p(y_{ 2 }|v_{ :2 }^{:D}=\nu _{ :2 }^{:D})p(v_{ 2 }^{:D}=\nu _{ 2 }^{:D}) }{ p(\varphi _{ 2 }^{:D}=\nu _{ 2 }^{:D}|\varphi _{ :1 }^{:D}=\nu _{ :1 }^{:D},y_{ :2 }) } \omega _{ 1 }^D(\nu _{ :1 }^{:D}) 
\end{align*}
which is equivalent to the weight update for the Particle Filter from Eq.~\ref{eq:seq_noise_weights}.  Since the denominator of the fraction in Eq.~\ref{eq:cpf_weight_recursion} is equal to the numerator of the previous time step, all the intermediate terms $p(y_{ 2 }|v_{ :2 }^{:d}=\nu _{ :2 }^{:d})$ with $d<D$ cancel each other out.

This is no surprise since we merely divided up the Particle Filter equations into smaller steps, but after iterating over all dimensions we will still obtain the same result for the weight. If there is no resampling in the CPF within a time step, it will yield the exact same result as the PF. Differences appear however when resampling is required within time steps, before all noise dimensions have been incorporated. The CPF will in that case discard unlikely samples early, which is not possible with the PF.

\subsection{Approximations}
Unfortunately, not all terms in Eq.~\ref{eq:cpf_initial_weight_recursion} and Eq.~\ref{eq:cpf_weight_recursion} are easy to compute. The terms $ \phi_1^d (\nu _{ :1 }^{ :d })$ are unproblematic, since the numerator is the process noise distribution, which is Gaussian, and the denominator is the proposal distribution, which we are free to choose. 

The terms of the form $p(y_{ 2 }|v_{ :2 }^{ :d })$ can however in most cases not be obtained in closed form, except of course for $p(y_{ 2 }|v_{ :2 }^{ :D })$, which is simply the likelihood and also occurs in the standard Particle Filter, see Eq.~\ref{eq:seq_noise_weights}. The terms for $d<D$ however require computing the integral
\begin{align}
p(y_{ 2 }|v_{ :2 }^{ :d })=\int _{ v_{ 2 }^{ d+1:D } } p(y_{ 2 }|v_{ :2 }^{:D})p(v_{ 2 }^{ d+1:D })\label{eq:partial_integral}
\end{align}
which in most cases cannot be done in closed form. Below we propose an approximation which is feasible for any dynamical system, it is however very important to note that all the approximate terms $p(y_{ 2 }|v_{ :2 }^{ :d })$ for $d<D$ cancel each other out, as shown above, and the full weight $ { \omega  } ^D_{ 2 }(\nu _{ :2 }^{:D})$ is exact.

Consequently, if no resampling in between full samples is required, i.e. we have enough samples to cover the $D$ dimensions of the state space, then the proposed algorithm is equivalent to the standard Particle Filter, no matter what approximation is used. If there is resampling in between full samples, then of course the approximation matters.

If the approximation provides a good indication as to whether the full sample will have a 
large weight, then we can expect the proposed method to work well. If however the approximation is poor, then resampling based on the partial weights can lead us to discard good candidates. To shed some light on these issues, we will compare the performance of the proposed method with the standard PF under different conditions in the experimental section.
\subsubsection*{Dirac Approximation}
There are many possible ways of approximating the integral from Eq.~\ref{eq:partial_integral}, the purpose is to have at least a rough idea of the likelihood of a sample before sampling all the dimensions. One option is to approximate the process noise $p(v_{ 2 }^{ d+1:D })$, which is a standard normal distribution, by a Dirac distribution centered at zero, the solution to the integral is then easy to obtain:
\begin{align*}
  { p } (y_{ 2 }|v_{ :2 }^{ :d })=p(y_{ 2 }|v_{ :2 }^{ :d },v_{ 2 }^{ d+1:D }=0).
\end{align*}
This very simple approximation worked well in most of our experiments and is applicable to any system model.

\section{Application to a Linear Gaussian System}
Let us consider a simple linear Gaussian system in order to illustrate the proposed method. The state of this system could of course be inferred optimally using a Kalman filter. Nonetheless it will provide us with some insight into the characteristics of the proposed filter.

\begin{figure}
  \centering
  \includegraphics[width=0.7\columnwidth]{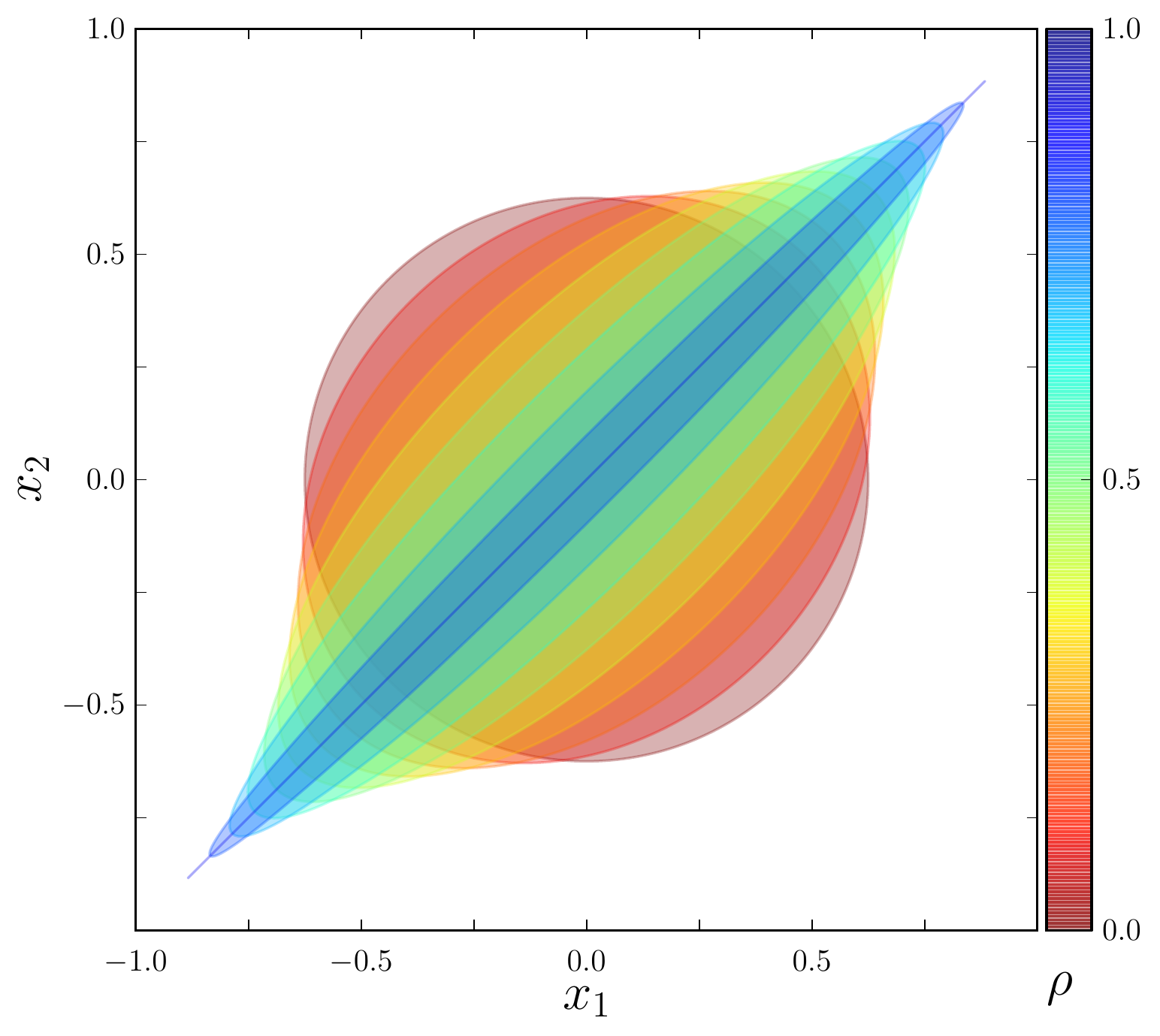}
  \caption{The covariance shape varies from a sphere ($\rho=0.0$, no correlation) to a line ($\rho=1.0$, maximum correlation).}
  \label{fig:toycovariances}
\end{figure}

For convenience, we write the process model in functional form
\begin{align*}
 { x }_{ 2 }=x_{ 1 }+ v_{ 2 }
\end{align*}
where $v_{2}$ is drawn from a standard normal distribution. The observation model is given as a distribution:
\begin{align*}
   p(y_{ 2 }|x_{ 2 })&=N(y_{ 2 }|x_{ 2 },  Q)
\end{align*}
For simplicity we will consider a system in which all the measurement variables have a variance equal to $1$. Furthermore we assume that the Pearson correlation between the different measurement dimensions $y^i$ and $y^j$ is equal to $\rho$ for all $i,j$. All the diagonal elements of $Q$ will thus be equal to one, and all the off diagonal elements equal to $\rho$. $Q$ has $D$ eigenvalues $\lambda_d$ where $\lambda_i = 1 - \rho$ for $i=1\cdots D-1$ and $\lambda_D = (D-1)x + 1$. This scaling is visualized for a two-dimensional system in Fig.~\ref{fig:toycovariances}.
Varying $\rho$ will allow us to investigate how the proposed method
behaves depending on how correlated different dimensions are. The
second parameter we will vary is the dimension $D$ of the system. 
In the following, we compare the standard Particle Filter with the
proposed exact and approximate Coordinate Particle Filter.

\subsection{Particle Filter}
We choose the proposal distribution according to our process model
\begin{align*}
p(\varphi _{ 2 }=\chi _{ 2 }|\varphi _{ :1 }=\chi _{ :1 },y_{ :2 })=N(\chi _{ 2 }|\chi _{ 1 }, I)
\end{align*}
the importance weights from Eq.~\ref{eq:seq_weight} then become
\begin{align*}
 \omega _{ 2 }(\chi_{ :2 })\propto N(y_{ 2 }|\chi _{ 2 },Q)\omega _{ 1 }(\chi _{ :1 })
\end{align*}
that is, we simply re-weight according to the observation model.
\subsection{Exact Coordinate Particle Filter}

First of all we will make the noise explicit. Our system is thus described by $p(y_{ 2 }|v_{ :2 })$. This distribution can be obtained from the process and observation model:
\begin{align*}
 p(y_{ 2 }|v_{ :2 }^{:D})=N(y_{ 2 }|\sum_t v_{ t }^{:D},Q).
\end{align*}
To be able to compute the partial weights, we have to solve the integral in Eq.~\ref{eq:partial_integral}.
For this very simple system, this integral can be solved analytically, which means that the partial weights in Eq.~\ref{eq:cpf_initial_weight_recursion} and Eq.~\ref{eq:cpf_weight_recursion} can be computed without any approximation. We will refer to the filter using these weights as the exact Coordinate Particle Filter.

\subsection{Approximate Coordinate Particle Filter}
In most cases we will not be able to solve Eq.~\ref{eq:partial_integral} analytically, we will thus apply the approximation method described above, where $p(v_{ 2 }^{ d+1:D })$ is approximated by a Dirac spike, for comparison. We refer to the resulting filter as the approximate Coordinate Particle Filter.
\subsection{Experimental Setup}
For each filtering algorithm we run 10 simulations per parameter set. At each time step we compute the root mean squared (RMS) error between the estimate by the filter and the true state. Finally, we compute the mean and variance of the RMS-error across the time steps and runs. We therefore end up with an error mean and variance per filter and parameter set, which can be interpreted as a Gaussian distribution over the error. To compare the CPF and the PF we use these two distributions to compute the probability of the error of the CPF being smaller than the one of the PF.
\subsection{Results}
\begin{figure}[h]
\centering
 \includegraphics[width=0.9\linewidth]{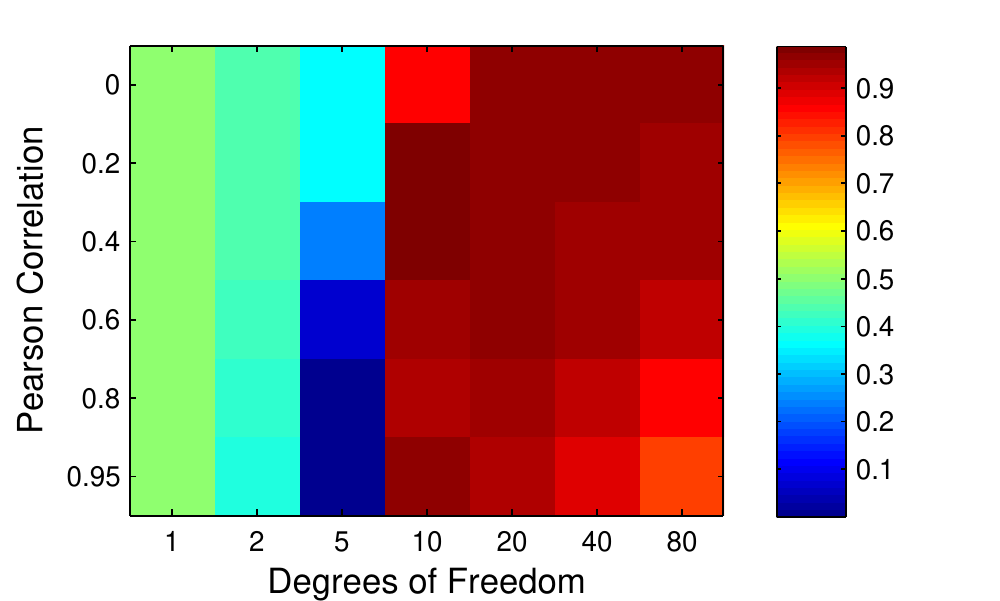}
    	\vspace{-0.25cm}
\caption{The probability of the error of the exact CPF being smaller than the error of the PF.}
 \label{fig:ecpf_vs_pf}
\end{figure}
In Fig.~\ref{fig:ecpf_vs_pf} we compare the exact CPF with the PF for different numbers of degrees of freedom and correlations. For dimension 1 and 2 both work well, there is thus not much difference, i.e. the probability of the exact CPF having a smaller error is close to 0.5. For the rest of the Fig. a clear trend can be seen: For high correlation and low dimensionality the standard Particle Filter performs better, and with increasing dimension and decreasing correlation the exact CPF gains more and more advantage.
\begin{figure}[h]
\centering
 \includegraphics[width=0.9\linewidth]{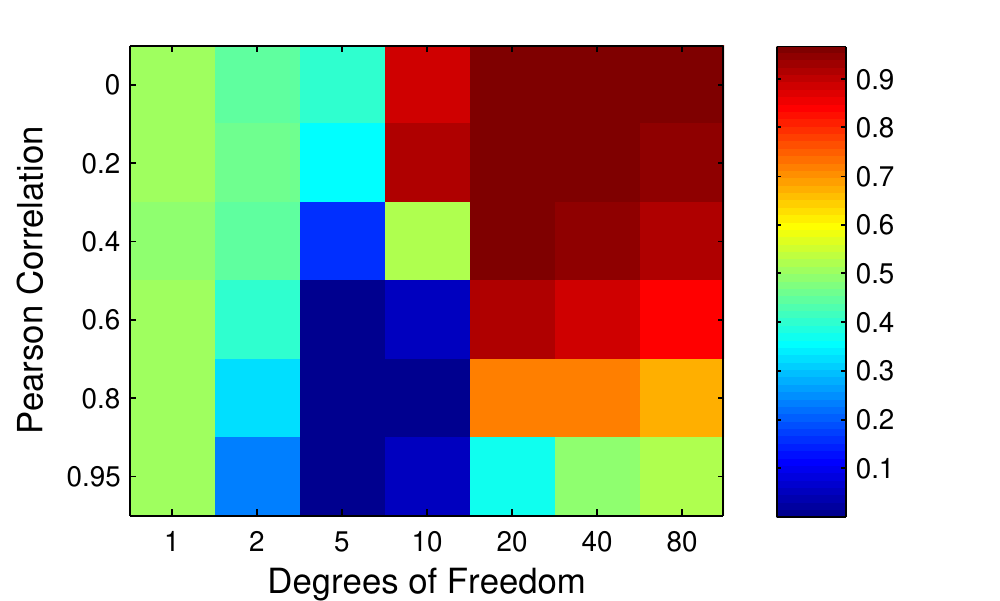}
    	\vspace{-0.25cm}
\caption{The probability of the error of the approximate CPF being smaller than the error of the PF.}
 \label{fig:acpf_vs_pf}
\end{figure}
A similar trend can be seen for the approximate CPF in Fig.~\ref{fig:acpf_vs_pf}, with the difference that the performance of the approximate method compares a little bit less favorably to the PF than the exact CPF. This trend is not surprising, since the more correlated the dimensions of a system, the harder it is to explore each dimension independently.

\begin{figure}[h]
\centering
 \includegraphics[width=0.9\linewidth]{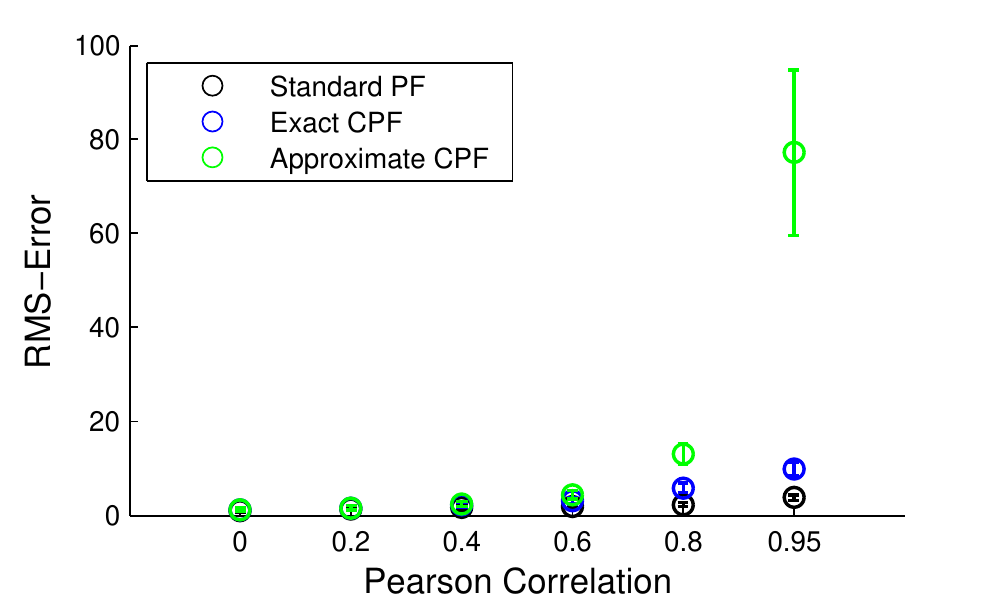}
   	\vspace{-0.25cm}
\caption{The error in function of the correlation for a fixed dimension $D=5$.}
 \label{fig:fixed_dim}
\end{figure}
In Fig.~\ref{fig:fixed_dim} we plot the error as a function of the correlation, this graph represents the same data as one column of the heat maps above. The performance of the approximate CPF degrades very quickly due to the crude approximation made.

\begin{figure}[h]
\centering
 \includegraphics[width=0.9\linewidth]{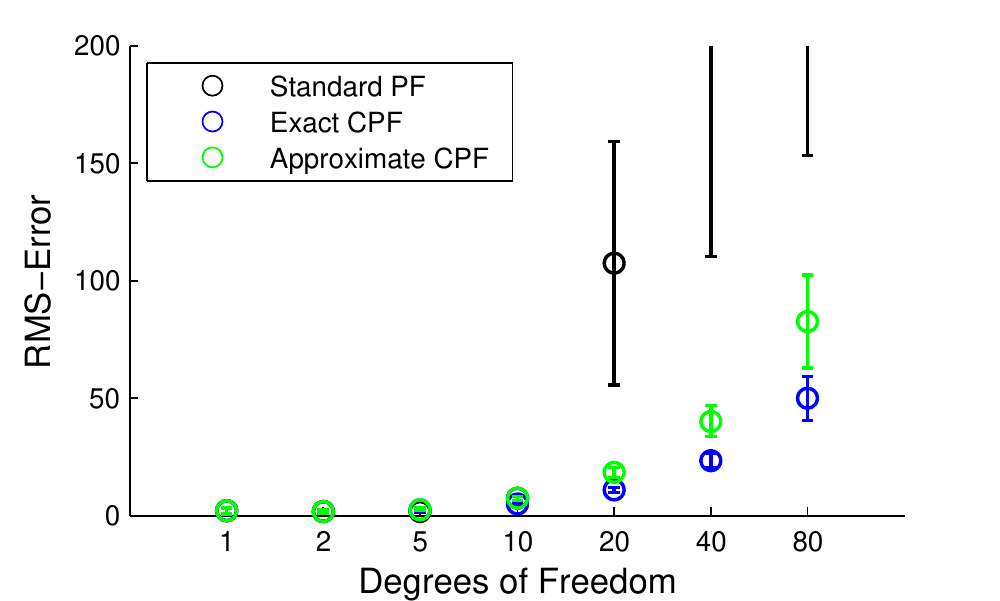}
    	\vspace{-0.25cm}
\caption{The error in function of the number of degrees of freedom for a fixed correlation of $0.4$. }
 \label{fig:fixed_corr}
\end{figure}
In Fig.~\ref{fig:fixed_corr} we plot the error as a function of the dimensionality. The performance of the Particle Filter degrades much quicker than the performance of the proposed methods.

The insight from this experiment is that in very high-dimensional systems the PF will perform poorly, whereas the proposed method can perform well if the dimensions are not too strongly correlated.

\section{Application to Tracking using Range Images}\label{sect:experiments}
We apply the Coordinate Particle Filter to 3D object tracking using an
RGB-D camera, as proposed in \cite{object_tracking}. The state consists of the 3D poses of the objects of interest and the observations are range images.
The observation model predicts the measurement to be in the proximity of the rendered 3D object model.

For evaluation we track several independently moving objects with the proposed Coordinate Particle Filter as well as with a standard Particle Filter. We could break the system down into low dimensional sub systems and track each object separately, which might work well since they are only weakly interacting. The objective of this experiment is however to understand whether the proposed method can exploit the structure of the system autonomously. This is important, since in different dynamical systems the algorithm might be able to exploit the structure in a way which is not as obvious as in this case. Furthermore, this setup allows for a straight forward analysis of the scalability with respect to the dimensionality of the state space, simply by increasing the number of objects.

\subsection{Experimental Setup}
We conducted experiments both in simulation and on real range data.
In both cases we continuously move boxes while tracking their poses.
To investigate scaling with the dimensionality of the state, we evaluate the algorithm with 1, 3, 6, 9 and 12 moving boxes.
Each box has 6 degrees of freedoms which need to be estimated.
Since both the Particle Filter and the proposed Coordinate Particle Filter are sampling based,
we run each experiment ten times on the same dataset in order to collect some statistics.

The Coordinate Particle Filter requires $N_{cpf}D$ evaluations of the likelihood per time step,
where $N_{cpf}$ is the number of particles and $D$ is the number of degrees of freedom.
The standard Particle Filter requires only $N_{pf}$ evaluations per time step,
where $N_{pf}$ is the number of particles.
For fair comparison we ensure that both filters use the same number of evaluations.
Therefore we set $N_{cpf} = \frac{N_{pf}}{D}$ and for all experiments we have $N_{pf}=2000$.

\subsection{Simulation}
\begin{figure}[h]
 \centering
 \includegraphics[width=0.9\linewidth]{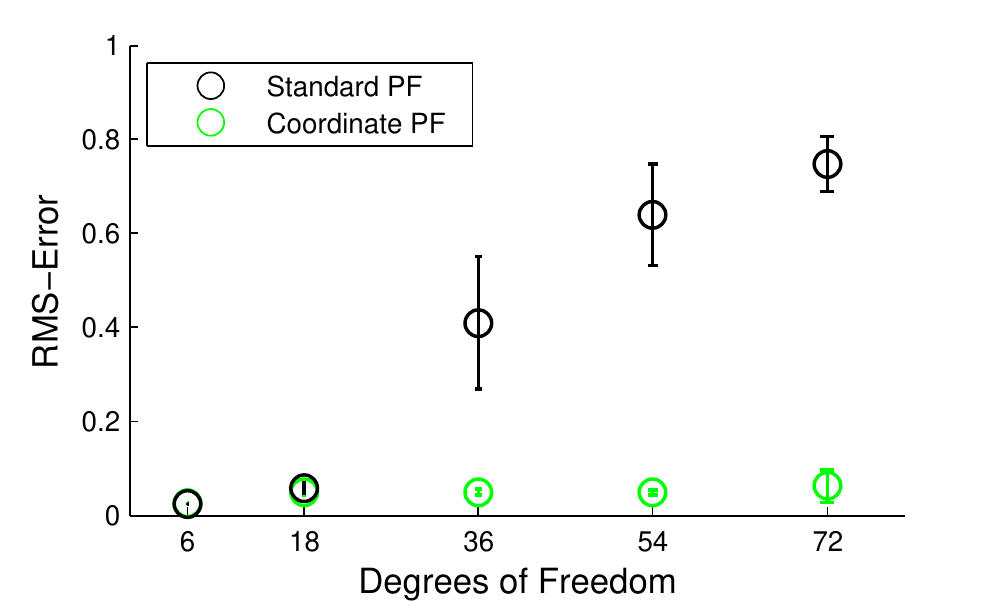}
  	\vspace{-0.25cm}
\caption{The proposed CPF is compared to the standard PF for different numbers of degrees of freedom on simulated range data.
The dimensionality of the state does not affect the performance of the proposed method, the standard PF however degrades rapidly. }
 \label{fig:evaluation_sim_data}
\end{figure}

The range data is generated from a simulation that exhibits
similar artifacts as a real RGB-D camera (occlusion boundaries,
quantization steps, perlin-like noise) \footnote{The source code is available at \url{https://github.com/jbohg/render_kinect}.}. 
For the evaluation of the accuracy we compute the root-mean-square error (RMSE) between the estimated pose and the true pose, averaged across all time steps. In Fig.~\ref{fig:evaluation_sim_data} we plot the mean and standard deviation of the RMSE across the ten runs.
For 6 and 18 dimensions there seems to be no significant difference between the CPF and the standard PF.
As expected, the standard PF starts degrading very rapidly for higher dimensional states, whereas the CPF seems to be able to exploit the fact that there are only few strong dependencies between the dimensions of this dynamical system.

%

\subsection{Real Data}
\begin{figure}[h]
 \centering
 \includegraphics[width=0.9\linewidth]{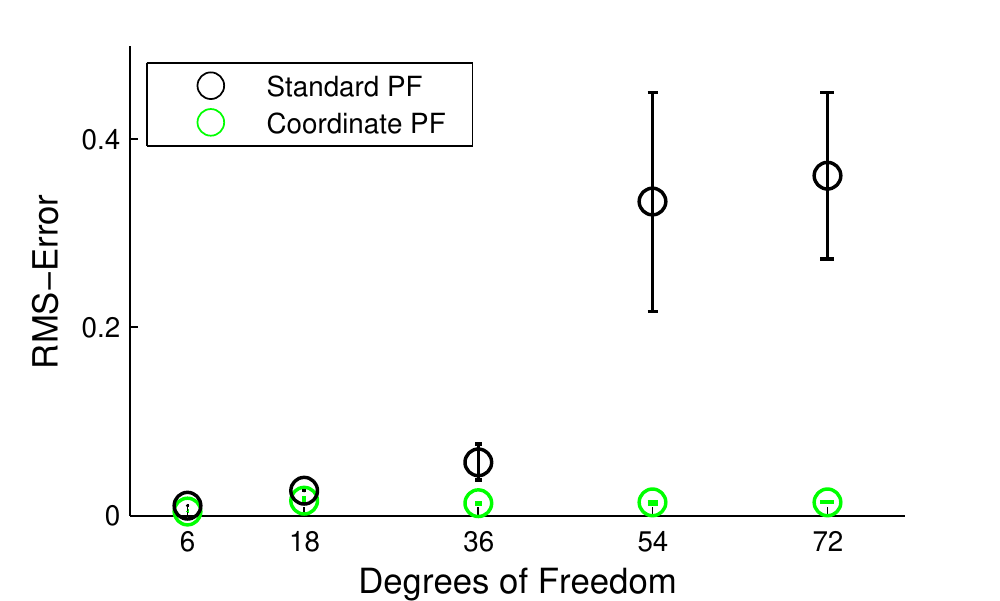}
  	\vspace{-0.25cm}
\caption{The CPF is compared to the standard PF for different numbers
  of degrees of freedom on \textbf{real range data} from an RGB-D camera.}
 \label{fig:evaluation_real_data}
\end{figure}

\begin{figure}[h]
  \centering
  	\includegraphics[width=\linewidth]{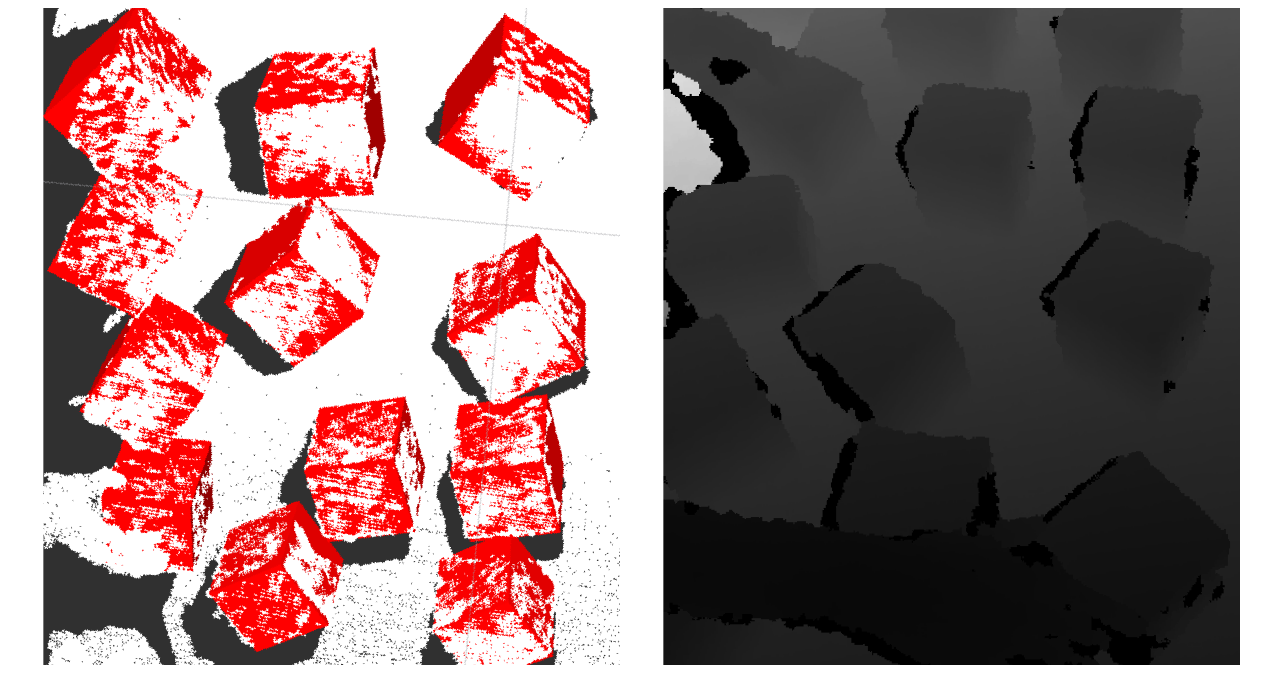}
 	 \caption{Real point cloud displayed on the left and
           the corresponding range image.}
  	\label{fig:kinect_real}
\end{figure}

To confirm the validity of the simulation based experiment, we run a
very similar experiment with real range images from an RGB-D camera (cf. Fig.~\ref{fig:kinect_real}).
We record a scene in which the corresponding number of boxes are manually moved.
In this setup no ground truth is available.
For the evaluation, we assume that the ground truth poses are equal to the mean poses across all individual runs on the same dataset.
Thus, in contrast to the simulation based experiments, the errors
plotted in Fig.~\ref{fig:evaluation_real_data} represent the
variance of the estimate. 
Nevertheless, Fig.~\ref{fig:evaluation_real_data} conveys the same
development as the simulation based experiments, shown in
Fig.~\ref{fig:evaluation_sim_data}.
This suggests, that the simulation experiments are indeed a good indicator for the performance on real data.

\section{Application to Manipulator Tracking}
\begin{figure}[h]
  \centering
  	\includegraphics[width=0.7
  	\linewidth]{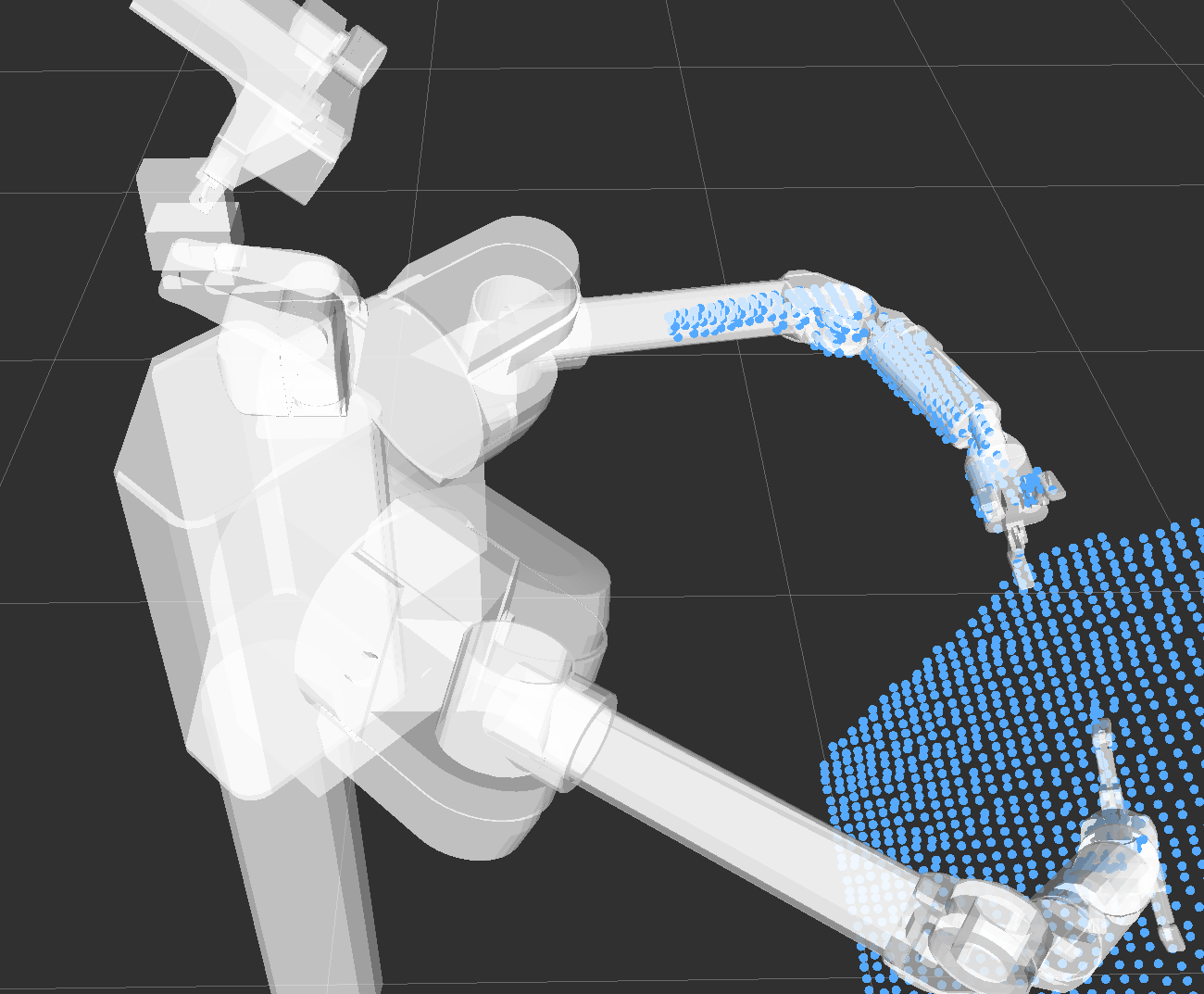}
 	 \caption{The robot model is represented in white, and the
           simulated range data in blue. The robot consists of two WAM arms with hands by Barrett~\cite{Righetti13}.}
  	\label{fig:roboto}
\end{figure}

In this section, we apply the proposed method to manipulator
tracking. We simultaneously track the 30 joint angles of the two arms, including the
fingers, of the robot in Fig.~\ref{fig:roboto}.  
The observation model is the same as described above, but now the
different objects are not moving freely anymore but are constrained
through the robot's kinematic chain. Unlike in the multi object
tracking example above, this system could not easily be broken down
into several weakly interacting subsystems which can be tracked
individually. 
\begin{figure}[h]
  \centering
  	\includegraphics[width=\linewidth]{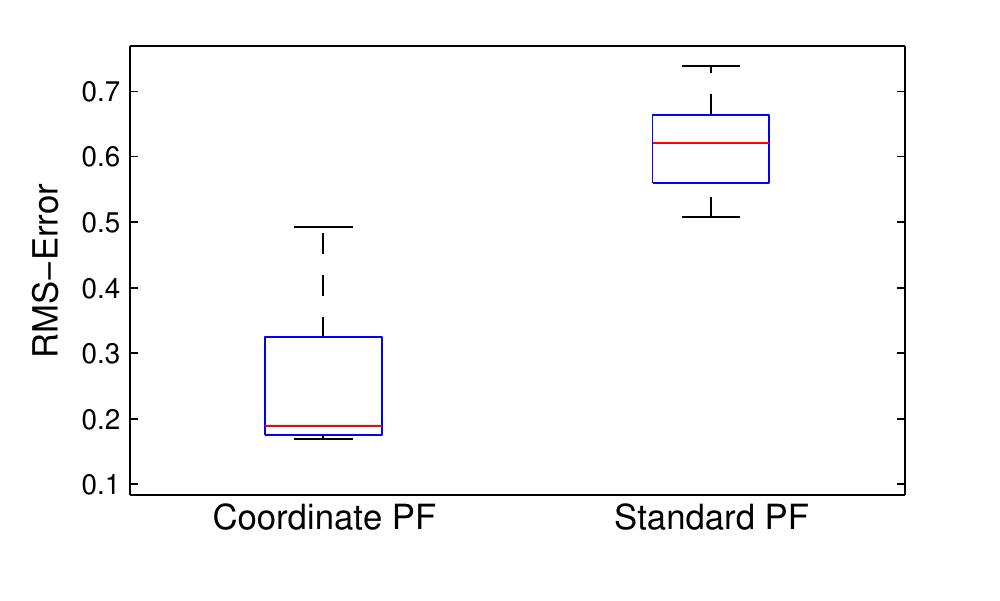}
  	\vspace{-1cm}
 	 \caption{The CPF is able to track the robot motion well despite the high dimensionality. The RMS error is computed between the estimated and ground-truth 30-dimensional joint configuration of the two robot arms.}
  	\label{fig:arm_tracking}
\end{figure}
Nevertheless, as we can see in Fig.~\ref{fig:arm_tracking}, the
proposed method outperforms the standard Particle Filter on
simulated range data. In the accompanying video, we show qualitative results for
tracking the real robot arm.

\section{CONCLUSIONS} \label{sect:conclusion}
To sufficiently cover the state space, the standard Particle Filter
requires the number of samples to grow exponentially in the
dimensionality of the state space. This precludes its usage in high
dimensional systems. We propose a method called the Coordinate
  Particle Filter (CPF), which  computes the approximate weights
recursively in both the time steps as well the degrees of freedom of the
state, as opposed to the standard PF which is recursive only in the
time steps. Therefore the CPF can
resample per dimension before even drawing a full state sample. This
prevents the particle weights from becoming degenerate even in high dimensional
state spaces.

Experimental results indicate that the proposed method is able to automatically 
exploit the structure of dynamical systems where not all dimensions are 
strongly dependent on each other. It can therefore perform well in very 
high-dimensional systems which are beyond the scope
of a standard Particle Filter.

An open question is what kind of dependency structures the proposed method is able to 
exploit. Another interesting direction
of future research is to investigate the importance of
the order in which dimensions are explored, and if an optimal order
could  be determined automatically.


%
%
%

 
{\scriptsize
\bibliographystyle{abbrvnat}
\bibliography{cpf}
}

\end{document}

%% file: hidden_markov.tex
\begin{tikzpicture}
   
   \node[latent]                      	  	(sta_pre) {$\sta_{\pre}$};
   \node[latent, right=of sta_pre]          	(sta_cur) {$\sta_{\cur}$};
   \node[latent, right=of sta_cur]          	(sta_nex) {$\sta_{\nex}$};
    
   \node[obs, below=of sta_pre]                  		(obs_pre) {$\obs_{\pre}$};
   \node[obs, below=of sta_cur, right=of obs_pre]         	(obs_cur) {$\obs_{\cur}$};
   \node[obs, below=of sta_nex, right=of obs_cur]         	(obs_nex) {$\obs_{\nex}$};
    
  \edge {sta_pre} {sta_cur} ; %
  \edge {sta_cur} {sta_nex} ; %
  
  \edge {sta_pre} {obs_pre}
  \edge {sta_cur} {obs_cur}
  \edge {sta_nex} {obs_nex}

\end{tikzpicture}

%% file: standard_particle_filter.tex
\begin{tikzpicture}
   \node[latent]                      	  	(sta_pre) {$\sta_{\pre}$};
   \node[latent, right=of sta_pre]          	(sta_cur) {$\sta_{\cur}$};
   \node[latent, right=of sta_cur]          	(sta_nex) {$\sta_{\nex}$};
    
   \node[obs, below=of sta_pre]                  		(obs_pre) {$\obs_{\pre}$};
   \node[obs, below=of sta_cur, right=of obs_pre]         	(obs_cur) {$\obs_{\cur}$};
   \node[obs, below=of sta_nex, right=of obs_cur]         	(obs_nex) {$\obs_{\nex}$};
   
   \node[latent, below=of obs_pre]                      	(sam_pre) {$\sam_{\pre}$};
   \node[latent, below=of obs_cur, right=of sam_pre]   	(sam_cur) {$\sam_{\cur}$};
   \node[latent, below=of obs_nex, right=of sam_cur]          	(sam_nex) {$\sam_{\nex}$};

  \edge {sta_pre} {sta_cur};
  \edge {sta_cur} {sta_nex};
  
  \edge {sta_pre} {obs_pre};
  \edge {sta_cur} {obs_cur};
  \edge {sta_nex} {obs_nex};
  
  \edge {obs_pre} {sam_pre};
  \edge {obs_pre} {sam_cur};
  \edge {obs_pre} {sam_nex};
  \edge {obs_cur} {sam_cur};
  \edge {obs_cur} {sam_nex};
  \edge {obs_nex} {sam_nex};
  
  \edge {sam_pre} {sam_cur};
  \edge[bend right] {sam_pre} {sam_nex};
  \edge {sam_cur} {sam_nex};
\end{tikzpicture}

%% file: noise_particle_filter.tex
\begin{tikzpicture}
  \node[latent]                      	  	(pno_pre) {$\pno_{\pre}$};
  \node[latent, right=of pno_pre]          	(pno_cur) {$\pno_{\cur}$};
  \node[latent, right=of pno_cur]          	(pno_nex) {$\pno_{\nex}$};
   
   \node[latent, below=of pno_pre]                     	(sta_pre) {$\sta_{\pre}$};
   \node[latent, below=of pno_cur, right=of sta_pre]   	(sta_cur) {$\sta_{\cur}$};
   \node[latent, below=of pno_nex, right=of sta_cur]   	(sta_nex) {$\sta_{\nex}$};
    
   \node[obs, below=of sta_pre]                  		(obs_pre) {$\obs_{\pre}$};
   \node[obs, below=of sta_cur, right=of obs_pre]         	(obs_cur) {$\obs_{\cur}$};
   \node[obs, below=of sta_nex, right=of obs_cur]         	(obs_nex) {$\obs_{\nex}$};
   
   \node[latent, below=of obs_pre]                      	(sam_pre) {$\sam_{\pre}$};
   \node[latent, below=of obs_cur, right=of sam_pre]   	(sam_cur) {$\sam_{\cur}$};
   \node[latent, below=of obs_nex, right=of sam_cur]          	(sam_nex) {$\sam_{\nex}$};

  \edge {sta_pre} {sta_cur};
  \edge {sta_cur} {sta_nex};

  \edge {pno_pre} {sta_pre};
  \edge {pno_cur} {sta_cur};
  \edge {pno_nex} {sta_nex};
  
  \edge {sta_pre} {obs_pre};
  \edge {sta_cur} {obs_cur};
  \edge {sta_nex} {obs_nex};
  
  \edge {obs_pre} {sam_pre};
  \edge {obs_pre} {sam_cur};
  \edge {obs_pre} {sam_nex};
  \edge {obs_cur} {sam_cur};
  \edge {obs_cur} {sam_nex};
  \edge {obs_nex} {sam_nex};
  
  \edge {sam_pre} {sam_cur};
  \edge[bend right] {sam_pre} {sam_nex};
  \edge {sam_cur} {sam_nex};
  
\end{tikzpicture}